\documentclass[lettersize,journal]{IEEEtran}

\usepackage{tabularx}
\usepackage{amsmath,amsfonts}
\usepackage{algorithmic}
\usepackage{algorithm}
\usepackage{array}
\usepackage{adjustbox}
\usepackage{graphicx}
\usepackage{subcaption}
\usepackage{booktabs}
\usepackage[table,xcdraw]{xcolor}
\usepackage{pifont}

\usepackage{multirow}
\usepackage{amssymb}
\usepackage{bm}
\usepackage{enumitem}
\usepackage{makecell}
\usepackage{hyperref}
\usepackage{soul}
\usepackage{stfloats}
\usepackage{url}
\usepackage{verbatim}
\usepackage{cite}
\usepackage{textcomp}
\usepackage[normalem]{ulem}
\useunder{\uline}{\ul}{}

\newcommand{\std}[1]{{\scriptsize$\pm$#1}}

\begin{document}

\title{EHHN: An Event-driven Heterogeneous Hypergraph Network for Object-Centric Next Activity Prediction}

\author{Jiaxing Wang, Kaitao Chen, Zhubin Han, Chenyu Hou, Bin Cao, Jing Fan and Ji Zhang
\thanks{*Bin Cao and Ji Zhang are corresponding authors.}
\thanks{Jiaxing Wang, Kaitao Chen, Zhubin Han, Chenyu Hou, Bin Cao, and Jing Fan are with the College of Computer Science and Technology, Zhejiang University of Technology, 310023, Hangzhou, China (e-mail: \{wjx, chenkaitao, hanzhubin, houcy, fanjing, bincao\}@zjut.edu.cn}
\thanks{Ji Zhang is with the University of Southern Queensland, Toowoomba, QLD 4350, Australia (e-mail: ji.zhang@unisq.edu.au)}
\thanks{Manuscript received XXX, 2026; revised XXX, 2026.}}

\markboth{IEEE TRANSACTIONS ON SERVICES COMPUTING,~Vol.~XX, No.~XX, XX~2026}
{Wang \MakeLowercase{\textit{et al.}}: EHHN: An Event-driven Heterogeneous Hypergraph Network for Object-Centric Next Activity Prediction}

\maketitle

\begin{abstract}
Next activity prediction helps service-oriented processes anticipate upcoming steps before delays, exceptions, or service-level risks occur. Most existing methods assume classical single-case event logs, whereas real service processes often involve events shared by multiple typed business objects. Object-centric event logs (OCELs) capture such interactions, but current predictors remain limited. Flattening-based approaches lose cross-object context, and native OCEL graph-based approaches encode multi-object events through pairwise relations. Existing models also do not jointly capture event-driven object state changes, inter-event timing, and global execution patterns. We propose EHHN, an Event-driven Heterogeneous Hypergraph Network for object-centric next activity prediction. EHHN represents each prediction prefix as a heterogeneous hypergraph, where event--object hyperedges bind retained co-participating objects and a lifecycle hyperedge groups the primary object's observed lifecycle events. Based on this representation, EHHN uses a dual-stream architecture in which a micro-spatial stream models event-driven object-state evolution and a macro-evolution stream captures temporal dynamics using retrieved global prototypes. The two streams are fused to predict the next activity. Experiments on four public OCEL benchmarks against nine baselines show that EHHN achieves the best accuracy and macro F1-score on all datasets, with improvements of up to 8.1 and 12.4 percentage points over the strongest baselines. Compared with the strongest OCEL-native graph baseline, EHHN also reduces peak GPU memory by up to $24\times$. Code is available at \url{https://github.com/chenkaitao1112/EHHN}.
\end{abstract}


\section{Introduction}\label{sec:introduction}

Next activity prediction plays an important role in predictive process monitoring (PPM) by enabling service providers to anticipate the upcoming behavior of running process instances~\cite{pasquadibisceglie2024prophet,wang2025mhg}. In an order-fulfillment service, for example, predicting the next activity of an order, such as delivery scheduling, return handling, or exception resolution, can guide resource allocation and trigger service-level agreement (SLA) risk alerts. Such prediction may depend not only on the order's observed lifecycle events, but also on events involving related objects such as items, packages, and invoices. Most existing predictors, however, are built on classical single-case event logs, where each event is associated with one process instance, naturally represented as an ordered event sequence~\cite{sun2024next,nguyen2024switch,zare2025innovative,oved2025snap,chen2025dualview,pasquadibisceglie2024prophet,wang2025higpp,wang2025mhg}. This single-case view cannot directly represent service events shared by multiple typed business objects or dependencies across related object lifecycles.

Object-centric event logs (OCELs) move beyond the single-case view by recording each event together with all typed objects it involves, thereby representing multi-object processes through many-to-many event-object relations~\cite{Ghahfarokhi2021ocel,Fioretto2025survey}. This richer representation also makes prediction more challenging: the model can no longer rely on a single activity sequence, because the next activity may depend on several related objects and the events they share. Existing OCEL-based predictors and closely related predictive monitoring methods mainly follow two directions. Flattening-based methods project the OCEL onto a selected object type and apply sequence models developed for classical logs, treating cross-object events from that single perspective~\cite{gherissi2022ocppm,rohrer2022predictive,galanti2023ocexpsys}. Native OCEL graph-based methods preserve more object-centric information by representing events, objects, and their relations as event-object graphs, which are encoded with graph neural networks for prediction~\cite{adams2022framework,adams2023preserving,smit2024hoeg,gherissi2025framework}.

Despite this progress, existing OCEL-based prediction methods still leave four limitations. First, flattening-based methods project multi-object events onto one selected perspective, losing co-participating objects and cross-object dependencies. Second, native OCEL graph-based methods usually represent each multi-object event through pairwise edges, weakening the joint binding among the event and all objects involved in the same occurrence. Third, existing models rarely distinguish the asymmetric roles of events and objects: events are transient triggers of state changes, while objects persist over time and accumulate these changes. Fourth, inter-event timing and global execution patterns are seldom modeled together, although they are important for distinguishing prefixes with similar local event-object structures but different timing or process variants. These limitations call for a representation that preserves multi-object event semantics and a model that jointly captures event-driven state evolution, inter-event timing, and global execution patterns.

To this end, we propose EHHN, an Event-driven Heterogeneous Hypergraph Network for object-centric next activity prediction. EHHN represents the observed prefix around a primary object as a heterogeneous hypergraph: event--object hyperedges bind each retained event with its retained co-participating objects, avoiding pairwise decomposition within the bounded prefix, while a lifecycle hyperedge groups the primary object's observed lifecycle events. On top of this representation, a dual-stream architecture processes the prefix. The micro-spatial stream treats events as transient triggers and objects as persistent states, modeling event-driven object-state transitions and lifecycle-constrained refinement to capture local object evolution. A heterogeneous interaction encoder then aligns event and object representations into a shared space. The macro-evolution stream first uses time-aware attention to model inter-event timing and then retrieves latent execution patterns from a global prototype memory to provide dataset-level guidance. The fused representation combines these local and global signals for next activity prediction.


\par The main contributions of this paper are as follows:
\begin{itemize}
    \item We introduce a bounded heterogeneous hypergraph representation for object-centric next activity prediction. For each prediction prefix, event--object hyperedges connect each retained event with its retained participating objects, preserving joint multi-object participation within the bounded prefix, while a lifecycle hyperedge groups the primary object's observed lifecycle events. This representation avoids single-perspective flattening and pairwise decomposition of retained event-object relations.

    \item We propose a micro-spatial encoder that models the asymmetric roles of events and objects in OCELs. The encoder treats events as transient sources of update signals and objects as persistent entities with evolving states, so object representations are updated by their incident events and further refined with the primary object's lifecycle context.

    \item We design a macro-evolution encoder for information not captured by local event-object updates. The encoder applies time-aware attention to the primary object's lifecycle sequence to model inter-event timing and retrieves prototypes from a global prototype memory to provide dataset-level execution-pattern guidance.

    \item We evaluate EHHN on four public OCEL benchmarks against nine representative baselines. EHHN achieves the best Accuracy and macro F1-score on all datasets, with improvements of up to 8.1 and 12.4 percentage points, respectively, and reduces peak GPU memory by up to $24\times$ over the strongest OCEL-native graph baseline.
\end{itemize}

\section{Related Work}\label{sec:related}
This section reviews two bodies of work most relevant to EHHN: PPM on classical single-case event logs, with an emphasis on next activity prediction, and PPM on OCELs.

\subsection{PPM on Classical Single-Case Event Logs}
\label{sec:related:classical}

Next activity prediction has been widely studied in PPM on classical single-case event logs~\cite{marquezchamorro2017survey}. Early studies relied on statistical, rule-based, probabilistic, and feature-engineered models~\cite{francescomarino2015clustering,bohmer2018probability,tama2019empirical}, while recent approaches are mainly sequence-based or graph-based. Sequence-based methods model each running case as an ordered event sequence and learn temporal dependencies with LSTM variants~\cite{tax2017predictive,camargo2019lstm,gunnarsson2023direct}, CNN-BiLSTM and attention-based models~\cite{sun2024next}, Transformer encoders~\cite{bukhsh2021processtransformer,ni2023predictive,jalayer2022ham,nguyen2024switch,zare2025innovative}, multi-view or knowledge-enhanced predictors~\cite{pasquadibisceglie2022multi,wang2023mitfm,chen2022multi,pasquadibisceglie2024jarvis,donadello2023knowledge}, and recent LLM-based methods~\cite{oved2025snap,Pasquadibisceglie2024lupin,Angelo2026Enhancing}. Graph-based methods improve structural expressiveness by constructing instance-, sequence-, homogeneous-, or heterogeneous-graph representations and encoding them with GNNs or graph-enhanced recurrent architectures~\cite{chiorrini2021exploiting,chiorrini2023multi,ramamaneiro2024embedding,deng2024sequentialgraphs,pasquadibisceglie2024prophet,wang2025higpp,wang2025mhg}.

Despite these advances, classical PPM methods assume that each event belongs to a single process instance, so they cannot natively model events shared by multiple typed objects or dependencies across object lifecycles. Applying them to OCELs requires object-perspective projection or pairwise graph conversion, losing part of the native multi-object semantics.

\subsection{PPM on OCELs}
\label{sec:related:ocel}

OCELs record each event together with all typed objects involved in it, enabling predictive monitoring beyond the single-case perspective. Existing OCEL predictors mainly follow two directions. Flattening-based methods convert OCELs into sequence or feature inputs and reuse classical predictive models~\cite{gherissi2022ocppm,rohrer2022predictive}. Although such methods show that OCEL-derived inputs can support prediction, the model no longer operates on the native multi-object event structure. Galanti~\emph{et~al.}~\cite{galanti2023ocexpsys} incorporate object interaction information, and later work uses GAT-based embeddings of object-centric directly-follows graphs as contextual information for an LSTM predictor~\cite{gherissi2024predictive}. These methods still rely on derived object-centric features or selected object-type perspectives.

Native OCEL graph-based approaches retain event-object topology more directly, but they mainly target predictive tasks other than next activity prediction. Adams~\emph{et~al.}~\cite{adams2022framework,adams2023preserving} study object-centric feature extraction and graph-preserving encodings for process-mining tasks. HOEG~\cite{smit2024hoeg} builds a heterogeneous event-object graph and applies an end-to-end GNN for remaining-time prediction. Gherissi~\emph{et~al.}~\cite{gherissi2025framework} propose a modular object-centric PPM framework based on graph-based process executions and evaluate it on remaining-time and event-count prediction.

Overall, existing OCEL predictors demonstrate the value of object-centric information, yet object-centric next activity prediction that preserves multi-object event binding remains underexplored. Existing methods either reduce OCELs to selected object perspectives or encode multi-object events through pairwise graph structures, and they have not jointly addressed multi-object event binding, event-object role asymmetry, inter-event timing, and dataset-level execution regularities. 

\section{Preliminaries and Problem Formulation}\label{sec:preliminaries}
This section introduces the notation and formal definitions used throughout the paper.

\subsection{Object-Centric Event Logs}
An Object-Centric Event Log (OCEL) records each business event together with all typed objects involved in it. Formally, an OCEL is a tuple
\[
\mathcal{L}=(E,O,OT,\pi_{\mathsf{act}},\pi_{\mathsf{time}},\pi_{\mathsf{obj}},\pi_{\mathsf{type}},\pi_{\mathsf{ea}},\pi_{\mathsf{oa}}),
\]
where $E$, $O$, and $OT$ are finite sets of events, objects, and object types; $\mathcal{A}$ is the activity label set; and $\mathcal{U}_T$ is the timestamp domain. The mappings $\pi_{\mathsf{act}}:E\to\mathcal{A}$ and $\pi_{\mathsf{time}}:E\to\mathcal{U}_T$ assign activity labels and timestamps; $\pi_{\mathsf{obj}}:E\to\mathcal{P}(O)\setminus\{\emptyset\}$ maps each event to a non-empty subset of participating objects, where $\mathcal{P}(O)$ denotes the set of all subsets of $O$; $\pi_{\mathsf{type}}:O\to OT$ assigns object types; and $\pi_{\mathsf{ea}},\pi_{\mathsf{oa}}$ assign attribute values to events and objects.

To define object traces consistently, we impose a deterministic order on events. Events are ordered by $\prec$ using the lexicographic key $(\pi_{\mathsf{time}}(e),\iota(e))$, where $\iota(e)$ is a deterministic event index used to break timestamp ties.

For an object $o\in O$, its \emph{object trace} is the sequence of events in which $o$ participates, sorted by $\prec$:
\[
\sigma(o)=\langle e_1,e_2,\dots,e_{n_o}\rangle,
\]
where each $e_i$ satisfies $o\in\pi_{\mathsf{obj}}(e_i)$, the sequence contains all such events, and $e_i\prec e_{i+1}$ for $1\leq i<n_o$. Since one event may appear in the traces of multiple objects, OCELs naturally encode cross-object dependencies that single-perspective projections cannot fully preserve.

\subsection{Primary-Object Perspective and Prediction Prefix}
A \emph{primary object type} $\tau^*\in OT$ specifies the object type whose lifecycle is monitored for prediction. For example, in a procure-to-pay process containing purchase orders, invoices, goods receipts, and suppliers, choosing purchase order as $\tau^*$ means predicting the next activity for each purchase order. A \emph{primary object} $o_{\mathsf{pri}}\in O$ is an object of this type, i.e., $\pi_{\mathsf{type}}(o_{\mathsf{pri}})=\tau^*$.

Given the ordered object trace $\sigma(o_{\mathsf{pri}})=\langle e_1,\dots,e_n\rangle$, each position $t\in\{1,\dots,n-1\}$ defines a prediction prefix for forecasting the activity of the next lifecycle event $e_{t+1}$. We denote by $\mathcal{P}_t(o_{\mathsf{pri}})$ the prediction prefix of primary object $o_{\mathsf{pri}}$ after observing its first $t$ lifecycle events:
\[
\mathcal{P}_t(o_{\mathsf{pri}})
=(o_{\mathsf{pri}},E_{\mathsf{hist}}^{(t)},O_{\mathsf{hist}}^{(t)},E_{\mathsf{ctx}}^{(t)}),
\]
with
\[
E_{\mathsf{hist}}^{(t)}=\{e_1,\dots,e_t\},\quad
O_{\mathsf{hist}}^{(t)}
=\Bigl(\bigcup_{e\in E_{\mathsf{hist}}^{(t)}}\pi_{\mathsf{obj}}(e)\Bigr)
\setminus\{o_{\mathsf{pri}}\},
\]
and
\[
E_{\mathsf{ctx}}^{(t)}
=\{e\in E\setminus E_{\mathsf{hist}}^{(t)}
\mid \pi_{\mathsf{time}}(e)<\pi_{\mathsf{time}}(e_t),\,
\pi_{\mathsf{obj}}(e)\cap O_{\mathsf{hist}}^{(t)}\neq\emptyset\}.
\]
Here, $E_{\mathsf{hist}}^{(t)}$ contains the observed lifecycle events of the primary object, $O_{\mathsf{hist}}^{(t)}$ contains the auxiliary objects that co-participate in these events, excluding the primary object itself, and $E_{\mathsf{ctx}}^{(t)}$ contains context events that occur before the timestamp of the prefix endpoint and involve at least one observed auxiliary object. This strict timestamp cutoff is applied only to context events, so non-lifecycle events with the same or later timestamp than $e_t$ are excluded from the prefix input. The prediction target is the next activity label $a_{t+1}=\pi_{\mathsf{act}}(e_{t+1})$, where $a_{t+1}\in\mathcal{A}$.

\subsection{Problem Formulation}
Given an OCEL $\mathcal{L}$, a primary object type $\tau^*$, and a prediction prefix $\mathcal{P}_t(o_{\mathsf{pri}})$, the object-centric next activity prediction problem is to learn a classifier $\Omega_\theta$ that maps the prefix to a probability distribution over activity labels in $\mathcal{A}$. The predicted activity label is $\hat{a}_{t+1}=\arg\max_{a'\in\mathcal{A}}[\Omega_\theta(\mathcal{P}_t(o_{\mathsf{pri}}))]_{a'}$, and the ground-truth label is $a_{t+1}=\pi_{\mathsf{act}}(e_{t+1})$.

\section{The EHHN Framework}\label{sec:method}

\begin{figure*}[!t]
\centering
\includegraphics[width=0.92\textwidth]{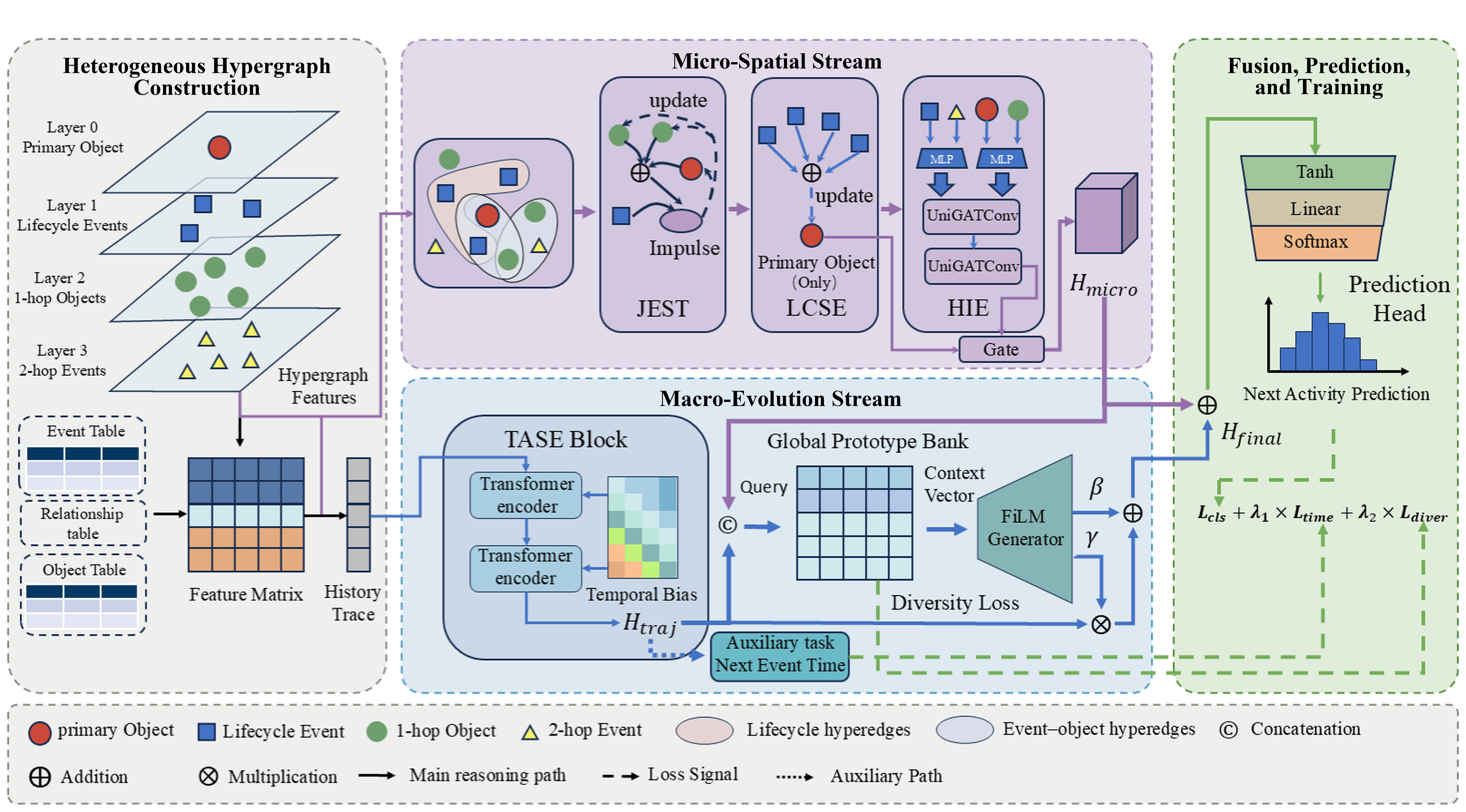}
\caption{Overview of the EHHN framework. A prediction prefix is converted into a heterogeneous prefix hypergraph with event--object and lifecycle hyperedges. The micro-spatial stream applies JEST, LCSE, and HIE to encode local event-driven object-state evolution, while the macro-evolution stream uses TASE and prototype memory to model temporal and global execution patterns. The two streams are fused for next activity prediction.}
\label{fig:framework}
\end{figure*}

EHHN is designed to preserve the native multi-object structure of OCEL prefixes while modeling both event-driven object-state evolution and temporal execution patterns. As shown in Fig.~\ref{fig:framework}, EHHN consists of four connected stages: heterogeneous hypergraph construction, micro-spatial stream, macro-evolution stream, and final fusion, prediction, and training. Given a prediction prefix, the construction stage builds a bounded heterogeneous prefix hypergraph with four node layers: the primary object, its observed lifecycle events, co-participating auxiliary objects, and context events associated with those auxiliary objects. The micro-spatial stream captures how observed events update persistent object states within the prefix hypergraph. The macro-evolution stream models inter-event timing along the primary-object lifecycle and retrieves latent execution patterns from a global prototype memory. Finally, the two representations are fused for next activity classification, and the model is trained end-to-end with classification, auxiliary time-prediction, and prototype-diversity objectives.

\subsection{Heterogeneous Hypergraph Construction}
\label{sec:method:hypergraph}

A prediction prefix in an OCEL contains not only the primary object's observed lifecycle, but also auxiliary objects that co-participate in its events and historical events associated with those objects. A sequential representation loses this cross-object context, while pairwise graph edges weaken the joint semantics of multi-object events. To preserve such structure within a controlled input size, EHHN constructs a bounded heterogeneous prefix hypergraph $\mathcal{H}=(\mathcal{V},\mathcal{R},\mathbf{X})$ for each prediction prefix. The construction consists of 2-hop neighborhood expansion, hyperedge construction, and node-feature encoding, which are detailed below.

\subsubsection{2-Hop Neighborhood Expansion}
Starting from a prediction prefix, EHHN first places the primary object and its observed lifecycle events into the first two layers:
$\mathcal{V}^{(0)}=\{o_{\mathsf{pri}}\}$ and $\mathcal{V}^{(1)}=E_{\mathsf{hist}}^{(t)}$.
It then expands to two outer layers: the auxiliary objects that co-participate in the observed lifecycle events,
$\mathcal{V}^{(2)}=O_{\mathsf{hist}}^{(t)}$, and the retained context events of those auxiliary objects:
\begin{equation}
\mathcal{V}^{(3)}=\{e\in E\setminus\mathcal{V}^{(1)}
\mid \pi_{\mathsf{time}}(e)<\pi_{\mathsf{time}}(e_t),\,
\pi_{\mathsf{obj}}(e)\cap\mathcal{V}^{(2)}\neq\emptyset\}.
\label{eq:two-hop-events}
\end{equation}
The endpoint $e_t$ is determined by the prediction position $t$, and the strict timestamp cutoff keeps only context events that occur before the prefix endpoint. The node set is $\mathcal{V}=\mathcal{V}^{(0)}\cup\mathcal{V}^{(1)}\cup\mathcal{V}^{(2)}\cup\mathcal{V}^{(3)}$, with object nodes $\mathcal{V}_O=\mathcal{V}^{(0)}\cup\mathcal{V}^{(2)}$ and event nodes $\mathcal{V}_E=\mathcal{V}^{(1)}\cup\mathcal{V}^{(3)}$. The ordered lifecycle sequence $\langle e_1,\dots,e_t\rangle$ is retained as the history trace of the primary object. This bounded 2-hop construction is chosen to retain the objects that directly interact with the primary object's observed lifecycle and the contextual event evidence associated with those objects. Expanding less would discard auxiliary-object context, whereas expanding further would introduce indirectly related objects and events that substantially increase graph size and may add weakly relevant noise.

\subsubsection{Hyperedge Types}
The bounded prefix contains two essential relational semantics: event-object participation and primary-object lifecycle membership. EHHN therefore uses two complementary hyperedge types to encode these relations. \emph{Event--object hyperedges} connect each retained event with the participating objects that are also retained in the prefix:
\begin{equation}
\mathcal{R}_{eo}
=\{\epsilon_e=\{e\}\cup(\pi_{\mathsf{obj}}(e)\cap\mathcal{V}_O)
\mid e\in\mathcal{V}_E\}.
\label{eq:event-object-hyperedges}
\end{equation}
Thus, each retained event is represented by one hyperedge rather than a set of pairwise edges, preserving the joint event-object binding within the bounded prefix. The intersection with $\mathcal{V}_O$ prevents context events in $\mathcal{V}^{(3)}$ from introducing additional object nodes outside the prefix. Consequently, a context-event hyperedge contains only the participating objects retained in the bounded prefix, which keeps the prefix size controlled and avoids introducing weakly related objects. 

The \emph{lifecycle hyperedge} groups the observed lifecycle events of the primary object:
\begin{equation}
\epsilon^{\mathsf{lc}}=\mathcal{V}^{(1)}.
\label{eq:lifecycle-hyperedge}
\end{equation}
This hyperedge captures lifecycle membership among the observed primary-object events, which is not expressed by individual event--object hyperedges alone. Since each prefix contains one primary-object lifecycle, $\mathcal{R}_{lc}=\{\epsilon^{\mathsf{lc}}\}$ contains a single lifecycle hyperedge. Together, the four node layers and two hyperedge types define the incidence structure of the heterogeneous prefix hypergraph $\mathcal{H}$, with $\mathcal{R}=\mathcal{R}_{eo}\cup\mathcal{R}_{lc}$.

\subsubsection{Node Features}
Each node in the heterogeneous prefix hypergraph $\mathcal{H}$ is assigned an initial feature vector derived from its object or event attributes. For object nodes, numerical attributes are min--max scaled and categorical attributes are one-hot encoded. Event nodes are processed analogously and are further augmented with temporal features. Specifically, each event node carries a cyclic timestamp encoding and two log-scaled relative-time features measured against the primary-object history trace: the elapsed time since the start of the trace and the elapsed time since the nearest preceding lifecycle event. Let $z_T(e)$ denote a timestamp component and $T$ its period: weekday with $T=7$, hour with $T=24$, and minute and second with $T=60$. For each component,
\begin{equation}
\boldsymbol{\phi}_T(e)=\bigl[\sin(2\pi z_T(e)/T),\;\cos(2\pi z_T(e)/T)\bigr].
\label{eq:cyclic-time}
\end{equation}
The encodings are concatenated over weekday, hour, minute, and second components. The resulting object and event features constitute the node-feature matrix $\mathbf{X}$.

Overall, this stage produces three outputs for the subsequent streams: hypergraph features formed by the retained hyperedges and incidence relations, the node-feature matrix $\mathbf{X}$, and the ordered history trace $\langle e_1,\dots,e_t\rangle$. The first two are passed to the micro-spatial stream, while the history trace is used by the macro-evolution stream.

\subsection{Micro-Spatial Stream: Event-Driven State Evolution}\label{sec:method:micro}

The micro-spatial stream is designed around an asymmetric view of OCEL nodes: events are transient triggers, while objects are persistent entities whose states evolve through event participation. Rather than treating events and objects as fully symmetric message-passing nodes, the stream updates object states through event-driven transitions. Using the hypergraph features and node-feature matrix produced by the construction stage, it outputs a representation $\mathbf{h}_{\mathsf{micro}}$ that captures how observed events have shaped the primary object's state.

As shown in Fig.~\ref{fig:framework}, the micro-spatial stream applies three operators in cascade. Joint Event-Driven State Transition (JEST) computes event-level update signals and propagates them to retained participating objects. Lifecycle-Constrained State Evolution (LCSE) injects lifecycle context into the primary-object state. Heterogeneous Interaction Encoder (HIE) aligns event and object representations into a shared space through hypergraph attention. A gated readout then combines the primary-object representation with a pooled retained-event representation to produce $\mathbf{h}_{\mathsf{micro}}$.

\subsubsection{JEST}
JEST is designed to preserve the joint participation of objects in a multi-object event while respecting the asymmetric roles of events and objects. Pairwise graph encoders split a multi-object event into independent links, whereas generic hypergraph operators often propagate messages symmetrically among incident nodes. In an OCEL prefix, however, an event is a transient occurrence that triggers state changes, while objects are persistent entities that accumulate the effects of events over their lifecycles. JEST therefore keeps each event as the source of an event-conditioned update signal and applies this signal to its retained participating objects. Let $\mathbf{h}_e$ and $\mathbf{h}_o$ denote the current representations of event node $e$ and object node $o$, respectively, and let $\mathcal{O}_e=\pi_{\mathsf{obj}}(e)\cap\mathcal{V}_O$ denote the participating objects of event $e$ retained in the prefix. The update consists of three phases.

\textit{Phase~1: Context aggregation.} For each event $e$, JEST aggregates the current representations of its retained participating objects to form a per-event context vector:
\begin{equation}
\mathbf{c}_e=\frac{1}{|\mathcal{O}_e|}\sum_{o\in\mathcal{O}_e}\psi_{\mathsf{ctx}}(\mathbf{h}_o),
\label{eq:jest-context}
\end{equation}
where $\psi_{\mathsf{ctx}}(\cdot)$ is a learnable object-context projection. The context vector summarizes the joint object state under which event $e$ occurs.

\textit{Phase~2: Update signal generation.} Given the context vector $\mathbf{c}_e$, JEST combines it with the event representation $\mathbf{h}_e$ to generate an event-level update signal:
\begin{equation}
\mathbf{i}_e=\mathrm{MLP}_{\mathsf{imp}}([\mathbf{h}_e\|\mathbf{c}_e]).
\label{eq:jest-signal}
\end{equation}
This signal encodes both the event attributes and the joint object state summarized by $\mathbf{c}_e$, and is shared by all objects in $\mathcal{O}_e$ as the impulse induced by event $e$.

\textit{Phase~3: Gated state update.} For each object $o\in\mathcal{V}_O$, let $\mathcal{N}(o)=\{e\in\mathcal{V}_E\mid o\in\mathcal{O}_e\}$ denote the retained incident events of $o$. The update signals from these events are summed and fed into a GRU cell:
\begin{equation}
\mathbf{h}_o^{\mathsf{spa}}=\mathrm{GRUCell}\!\Bigl(\textstyle\sum_{e\in\mathcal{N}(o)}\mathbf{i}_e,\;\mathbf{h}_o\Bigr),
\label{eq:jest-update}
\end{equation}
where $\mathbf{h}_o^{\mathsf{spa}}$ is the spatially updated object state. Although an event-level signal $\mathbf{i}_e$ is shared by all retained objects participating in event $e$, object-specific differentiation arises because each object accumulates signals from its own incident-event set $\mathcal{N}(o)$ and updates a per-object hidden state through the GRU gate. The summation models the cumulative effect of multiple retained events, while the GRU gate controls how much of the accumulated event-driven signal is absorbed and how much prior object state is retained.

\subsubsection{LCSE}
JEST captures event-triggered object updates but does not explicitly inject the primary object's lifecycle context. Two primary objects may have similar recent events but different accumulated histories, leading to different next activities. LCSE addresses this by using the lifecycle hyperedge together with the primary-object index to aggregate the observed lifecycle events into a lifecycle profile $\mathbf{p}_{o_{\mathsf{pri}}}$, and then incorporating this profile into the primary-object state through a GRU cell while leaving auxiliary objects unchanged:
\begin{equation}
\mathbf{p}_{o_{\mathsf{pri}}}=\frac{1}{|\mathcal{V}^{(1)}|}\sum_{e\in\mathcal{V}^{(1)}}\psi_{\mathsf{lc}}(\mathbf{h}_e),
\label{eq:lcse-profile}
\end{equation}
where $\psi_{\mathsf{lc}}(\cdot)$ is a learnable lifecycle-event projection.
\begin{equation}
\mathbf{h}_o^{\mathsf{lc}}=
\begin{cases}
\mathrm{GRUCell}(\mathbf{p}_{o_{\mathsf{pri}}},\mathbf{h}_o^{\mathsf{spa}}), & \text{if }o=o_{\mathsf{pri}},\\
\mathbf{h}_o^{\mathsf{spa}}, & \text{otherwise,}
\end{cases}
\label{eq:lcse-update}
\end{equation}
for each object node $o\in\mathcal{V}_O$. Only the primary object receives the lifecycle profile because the prediction target is its next activity. Auxiliary objects keep their JEST-updated states and serve as contextual evidence.

\subsubsection{HIE}
After JEST and LCSE, object states encode event-driven updates and lifecycle context, while event nodes retain activity, attribute, and temporal information. Because these object-state and event-node representations are complementary but heterogeneous, HIE first projects them into a shared $D$-dimensional space using separate MLPs for event nodes and object nodes. It then applies two residual UniGAT hypergraph attention layers~\cite{Huang2021UniGNNAU} to refine node representations over the merged incidence structure containing retained event--object hyperedges and the lifecycle hyperedge. The UniGAT layers do not use separate type-specific hyperedge parameters or hyperedge features. Relation-specific inductive biases are provided before this shared refinement: JEST models event-object participation through event-driven updates, LCSE injects lifecycle context, and event/object-specific projections handle node heterogeneity.

For prediction, the readout combines the primary-object state with event-side context from the retained prefix. We extract the refined representation of the primary object as $\mathbf{h}_{\mathsf{micro}}^o$ and obtain $\mathbf{h}_{\mathsf{micro}}^e$ by mean-pooling selected retained event-node representations. In implementation, this pooling uses up to the last three retained event nodes of each prefix. A learned gate combines these two representations:
\begin{align}
\mathbf{g}&=\sigma(\mathbf{W}_{\mathsf{ro}}[\mathbf{h}_{\mathsf{micro}}^o\|\mathbf{h}_{\mathsf{micro}}^e]+\mathbf{b}_{\mathsf{ro}}), \\
\mathbf{h}_{\mathsf{micro}}&=\mathbf{g}\odot\mathbf{h}_{\mathsf{micro}}^o+(1-\mathbf{g})\odot\mathbf{h}_{\mathsf{micro}}^e,
\label{eq:readout-fusion}
\end{align}
where $\odot$ denotes element-wise multiplication. The resulting $\mathbf{h}_{\mathsf{micro}}$ summarizes the primary-object state and the retained event-side context.

\subsection{Macro-Evolution Stream: Temporal Guidance and Global Modulation}\label{sec:method:macro}

The micro-spatial stream focuses on local event-object interactions and lifecycle-aware object states. To complement it, the macro-evolution stream captures two sequence-level signals that are not explicitly modeled by local structural updates. First, it models inter-event timing, since two executions with the same activity sequence may evolve over very different time scales. Second, it captures global execution patterns, since similar local prefixes may belong to different process variants. The macro-evolution stream implements these two functions through Time-Aware State Evolution (TASE), which takes the primary object's lifecycle event feature sequence as input and injects inter-event timing into attention, and a global prototype memory, which retrieves process prototypes that summarize dataset-level execution patterns.

\subsubsection{TASE}
TASE extends a standard Transformer encoder to model inter-event timing over the primary object's lifecycle events in chronological order. Its input is the preprocessed feature sequence of the observed lifecycle events, rather than the HIE-refined event representations from the micro-spatial stream. Let $\mathbf{x}_{e_i}$ denote the feature vector of lifecycle event $e_i$ after node-feature preprocessing and projection to dimension $D$. The TASE input is $\langle \mathbf{x}_{e_1},\dots,\mathbf{x}_{e_t}\rangle$, padded and masked when needed. Unlike standard Transformers that assign attention weights based only on feature similarity, TASE adds a logarithmic time-decay bias $\mathbf{B}_{\mathsf{time}}$ to the attention scores:

\begin{align}
\mathrm{Attn}(\mathbf{Q},\mathbf{K},\mathbf{V})
&=\mathrm{Softmax}\!\left(\frac{\mathbf{Q}\mathbf{K}^{\top}}{\sqrt{d}}+\mathbf{B}_{\mathsf{time}}\right)\mathbf{V},
\label{eq:tase-attn}\\
\mathbf{B}_{\mathsf{time}}[i,j]
&=-|w|\cdot\log(1+\Delta t_{ij})+b.
\label{eq:tase-bias}
\end{align}
The distance $\Delta t_{ij}$ is derived from relative lifecycle gaps. Let $g_i$ be the raw gap in seconds between lifecycle event $e_i$ and its preceding lifecycle event, with $g_1=0$. We use $\delta_i=(\log(1+g_i)-\mu_g)/(\sigma_g+\epsilon)$, where $\mu_g$ and $\sigma_g$ are computed from the training partition and $\epsilon$ is a small constant, set $s_i=\sum_{\ell=1}^{i}\delta_\ell$, and define $\Delta t_{ij}=|s_i-s_j|$. The scalars $w,b\in\mathbb{R}$ are learnable, and no additional clipping is applied. TASE stacks two such attention layers with four attention heads. As $\Delta t_{ij}$ increases, the decay term becomes more negative and reduces attention between temporally distant events. The logarithmic form compresses large temporal gaps, matching the irregular timing of business processes. The output at the last observed lifecycle position is used as the trajectory representation $\mathbf{h}_{\mathsf{traj}}\in\mathbb{R}^D$, which summarizes the primary object's temporal execution and serves as input to the global prototype memory. During training, an auxiliary regression head attached to $\mathbf{h}_{\mathsf{traj}}$ predicts the time gap to the next event. This auxiliary task uses the Smooth L1 loss, which is robust to large regression errors, and provides additional supervision for temporal representation learning.

\subsubsection{Global Prototype Memory and FiLM Modulation}
The global prototype memory complements TASE by injecting global execution patterns into the trajectory representation. We maintain a learnable prototype matrix $\mathbf{P}\in\mathbb{R}^{K\times D}$, where each row represents a latent execution pattern learned from the training data. A query vector is formed by projecting the concatenation of the micro-spatial representation and the trajectory representation:
\begin{equation}
\mathbf{q}=\mathrm{LayerNorm}(\mathbf{W}_q[\mathbf{h}_{\mathsf{micro}}\|\mathbf{h}_{\mathsf{traj}}]+\mathbf{b}_q),\quad\mathbf{W}_q\in\mathbb{R}^{D\times2D}.
\label{eq:proto-query}
\end{equation}
Before retrieval, the query and prototypes are $\ell_2$-normalized as $\widetilde{\mathbf{q}}=\mathbf{q}/\|\mathbf{q}\|_2$ and $\widetilde{\mathbf{P}}_i=\mathbf{P}_i/\|\mathbf{P}_i\|_2$ for the $i$-th prototype, and $\widetilde{\mathbf{P}}$ stacks all row-wise normalized prototypes. Let $\rho_i=\widetilde{\mathbf{q}}\widetilde{\mathbf{P}}_i^\top/\tau$, where $\tau$ is the temperature, and let $\mathcal{S}$ be the index set of the top-$k_p$ entries in $\{\rho_i\}_{i=1}^K$. The retrieval weights are
\begin{equation}
w_i=
\begin{cases}
\frac{\exp(\rho_i)}{\sum_{j\in\mathcal{S}}\exp(\rho_j)}, & i\in\mathcal{S},\\
0, & i\notin\mathcal{S},
\end{cases}
\quad
\mathbf{c}_{\mathsf{global}}=\sum_{i\in\mathcal{S}}w_i\widetilde{\mathbf{P}}_i .
\label{eq:proto-retrieval}
\end{equation}
Here, $\mathbf{c}_{\mathsf{global}}$ is the retrieved global context. The global context modulates $\mathbf{h}_{\mathsf{traj}}$ through feature-wise linear modulation (FiLM)~\cite{brockschmidt2020gnnfilm}:
\begin{equation}
\boldsymbol{\gamma}=\mathbf{1}+\alpha\tanh(\mathrm{MLP}_\gamma(\mathbf{c}_{\mathsf{global}})),\quad\boldsymbol{\beta}=\alpha\tanh(\mathrm{MLP}_\beta(\mathbf{c}_{\mathsf{global}})),
\label{eq:film-params}
\end{equation}
where $\alpha$ bounds the modulation amplitude. The modulated trajectory representation is $\boldsymbol{\gamma}\odot\mathbf{h}_{\mathsf{traj}}+\boldsymbol{\beta}$. The final FiLM layers are zero-initialized so that the model starts from an unmodulated trajectory representation and gradually learns how much global context to use.

\subsection{Fusion, Prediction, and Training}\label{sec:method:fusion}

\textit{Stream fusion.}
The FiLM-modulated trajectory representation is added to the micro-spatial representation:
\begin{equation}
\mathbf{h}_{\mathsf{final}}=\mathbf{h}_{\mathsf{micro}}+(\boldsymbol{\gamma}\odot\mathbf{h}_{\mathsf{traj}}+\boldsymbol{\beta}).
\label{eq:fusion}
\end{equation}
This additive fusion treats the micro-spatial representation as the base signal and lets the macro-evolution stream provide a learned trajectory correction. With zero-initialized FiLM layers, prototype-based modulation is inactive at initialization and is learned gradually during training.

\textit{Prediction head.}
The fused representation is mapped to the activity-label space by a two-layer prediction head:
\begin{equation}
\hat{\mathbf{y}}=\mathrm{Softmax}(\mathbf{W}_{\mathsf{pred}}\tanh(\mathbf{W}_{\mathsf{fuse}}\mathbf{h}_{\mathsf{final}}+\mathbf{b}_{\mathsf{fuse}})).
\label{eq:pred-head}
\end{equation}
Here, $\hat{\mathbf{y}}$ is the predicted probability distribution over activity labels. This prediction head instantiates the classifier $\Omega_\theta$ defined in Section~\ref{sec:preliminaries}.

\textit{Tri-objective training loss.}
The model is trained with a main classification loss, an auxiliary time-prediction loss, and a prototype-diversity regularizer:
\begin{equation}
\mathcal{J}=\mathcal{L}_{\mathsf{CE}}+\lambda_1(s)\mathcal{L}_{\mathsf{aux}}+\lambda_2\mathcal{L}_{\mathsf{div}}.
\label{eq:total-loss}
\end{equation}
where $\mathcal{L}_{\mathsf{CE}}$ is the cross-entropy loss for next activity prediction, $\mathcal{L}_{\mathsf{aux}}$ is the Smooth L1 loss for predicting the time gap to the next event, and
\begin{equation}
\mathcal{L}_{\mathsf{div}}=\|\widetilde{\mathbf{P}}\widetilde{\mathbf{P}}^\top-\mathbf{I}_K\|_F^2
\label{eq:proto-div}
\end{equation}
encourages different prototype vectors to be orthogonal and prevents prototype collapse. This term is used only when the global prototype memory is retained; it is omitted in ablation variants without prototypes. The weight $\lambda_1(s)$ follows a linear warmup over training step $s$ so the activity classifier stabilizes before time-prediction supervision takes effect.

\section{Experiments and Evaluation}\label{sec:experiments}
This section presents the experimental setup and empirical evaluation of EHHN. 
The evaluation addresses five research questions: overall effectiveness against representative baselines (RQ1), component contributions through ablation studies (RQ2), training time, inference latency, and GPU memory cost (RQ3), sensitivity to the global prototype count $K$ (RQ4), and prototype interpretability and robustness to attribute and event-feature noise (RQ5).

\subsection{Experimental Setup}\label{subsec:setup}

\subsubsection{Datasets}
We evaluate EHHN on four publicly available OCEL benchmarks. \textbf{BPI~2017}\footnote{\url{https://github.com/ocpm/ocpa/blob/main/sample_logs/csv/BPI2017.zip}} is an object-centric version of the widely used Dutch loan application process log and serves as the largest benchmark in our study. \textbf{OTC}\footnote{\url{https://github.com/niklasadams/PreservingOCStructures/blob/main/orders.zip}} is an order-management OCEL involving customers, orders, items, products, and packages. \textbf{Intermediate}\footnote{\url{https://github.com/ocpm/ocpa/blob/main/sample_logs/jsonocel/intermediate.jsonocel}} is an application-offer OCEL involving application and offer objects. \textbf{P2P}\footnote{\url{https://github.com/ocpm/ocpa/blob/main/sample_logs/jsonocel/p2p-2023.jsonocel}} is a procure-to-pay OCEL involving purchase requisitions, purchase orders, quotations, goods receipts, invoice receipts, payments, and materials.

Table~\ref{tab:datasets} summarizes the basic statistics and complexity of the four OCEL benchmarks. The basic statistics report event, object, activity, primary-object, and prefix counts. Let $O^*$ be the set of primary objects, i.e., $O^*=\{o\in O\mid \pi_{\mathsf{type}}(o)=\tau^*\}$, and let $E^*$ be the set of events involving at least one primary object, i.e., $E^*=\{e\in E\mid \pi_{\mathsf{obj}}(e)\cap O^*\neq\emptyset\}$.

For structural complexity, objects per event is the average event arity, $\frac{1}{|E|}\sum_{e\in E}|\pi_{\mathsf{obj}}(e)|$. Non-primary objects per event in $E^*$ is $\frac{1}{|E^*|}\sum_{e\in E^*}|\{o\in\pi_{\mathsf{obj}}(e)\mid \pi_{\mathsf{type}}(o)\neq\tau^*\}|$. Event-size entropy measures the diversity of event arities and is defined as $H_{\mathsf{size}}=-\sum_k p_d(k)\ln p_d(k)$, where $p_d(k)=|\{e\in E\mid |\pi_{\mathsf{obj}}(e)|=k\}|/|E|$. For object-perspective complexity, object-type co-occurrence entropy measures the diversity of event-level object-type signatures: $H_{\mathsf{type}}=-\sum_c p_{\mathsf{type}}(c)\ln p_{\mathsf{type}}(c)$, where $c(e)=\{\pi_{\mathsf{type}}(o)\mid o\in\pi_{\mathsf{obj}}(e)\}$ and $p_{\mathsf{type}}(c)=|\{e\in E\mid c(e)=c\}|/|E|$. Primary-object participation is the fraction of events involving primary objects, $\rho_{\mathsf{pri}}=|E^*|/|E|$.

For temporal complexity, let $\Delta t_o=(\Delta t_{o,1},\dots,\Delta t_{o,n_o-1})$ be the sequence of timestamp gaps between consecutive events in object trace $\sigma(o)$. Temporal gap variability is the mean object-level coefficient of variation, $\frac{1}{|O_{\Delta}|}\sum_{o\in O_{\Delta}}\mathrm{std}(\Delta t_o)/\mathrm{mean}(\Delta t_o)$, where $O_{\Delta}$ contains objects with at least two gaps. Let $r(e)$ be the rank of event $e$ in the deterministic global event order $\prec$. Global event-order gap is $\frac{\sum_{o\in O^*}\sum_{i=1}^{|\sigma(o)|-1}(r(e_{i+1})-r(e_i)-1)}{\sum_{o\in O^*}(|\sigma(o)|-1)}$, where $\sigma(o)=\langle e_1,\dots,e_{|\sigma(o)|}\rangle$ is the ordered trace of primary object $o$.

Overall, OTC is the most structurally complex benchmark, with the highest objects per event, non-primary objects per primary-object event, and event-size entropy. BPI~2017 is the largest benchmark and has the largest global event-order gap, indicating long-range separation between consecutive primary-object lifecycle events. P2P shows the strongest object-perspective complexity, with the highest object-type co-occurrence entropy and the lowest primary-object participation. Intermediate does not dominate any single complexity dimension and provides a relatively balanced benchmark.

\begin{table}[t]
\centering
\caption{Basic statistics and complexity of the benchmarks.}
\label{tab:datasets}
\footnotesize
\setlength{\tabcolsep}{3.5pt}
\renewcommand{\arraystretch}{1.08}
\begin{tabular}{lrrrr}
\toprule
Measure & OTC & BPI~2017 & Interm. & P2P \\
\midrule
\multicolumn{5}{l}{\emph{Basic statistics}} \\
\#Event instances & 22{,}367 & 393{,}931 & 20{,}641 & 8{,}320 \\
\#Object instances & 11{,}522 & 74{,}504 & 3{,}686 & 5{,}379 \\
\#Activity labels & 11 & 21 & 20 & 10 \\
\#Primary-object instances & 8{,}159 & 31{,}509 & 1{,}000 & 896 \\
\#Prediction prefixes & 56{,}758 & 297{,}385 & 15{,}668 & 4{,}302 \\
\midrule
\multicolumn{5}{l}{\emph{Structural complexity}} \\
Objects per event & 8.15 & 1.35 & 1.33 & 2.45 \\
Non-primary objs. per primary event & 5.25 & 0.41 & 0.41 & 1.43 \\
Event-size entropy & 2.08 & 0.68 & 0.67 & 0.70 \\
\midrule
\multicolumn{5}{l}{\emph{Object-perspective complexity}} \\
Object-type co-occurrence entropy & 0.49 & 1.01 & 1.03 & 1.64 \\
Primary-object participation & 1.00 & 0.83 & 0.81 & 0.57 \\
\midrule
\multicolumn{5}{l}{\emph{Temporal complexity}} \\
Temporal gap variability & 1.33 & 1.88 & 1.66 & 0.95 \\
Global event-order gap & 131.78 & 2484.59 & 101.00 & 116.44 \\
\bottomrule
\end{tabular}
\end{table}

\subsubsection{Baselines}
We compare EHHN against nine representative baselines. \emph{(i) Flattening/adapted classical methods:} \textbf{Dual-View}~\cite{chen2025dualview} and \textbf{PROPHET}~\cite{pasquadibisceglie2024prophet} are state-of-the-art predictors adapted to OCEL by treating each object type as a separate perspective; \textbf{Flat-LSTM}~\cite{tax2017predictive} applies a standard LSTM over flattened sequences; \textbf{GAT+LSTM}~\cite{gherissi2024predictive} combines a GAT-based snapshot encoder with an LSTM decoder. \emph{(ii) Native OCEL methods:} \textbf{HOEG}~\cite{smit2024hoeg} builds a heterogeneous event-object graph with GNN; \textbf{GCN}~\cite{kipf2017semi}, \textbf{GAT}~\cite{Velickovic2017GraphAN}, \textbf{GraphTransformer}~\cite{Yun2019GraphTN}, and \textbf{Logistic Regression}~\cite{Pedregosa2011ScikitlearnML} follow a two-stage OCEL graph-embedding pipeline in which the prefix is first encoded into a graph-level representation and a downstream predictor is applied. Every baseline is tuned on the validation set within the same hyperparameter search budget. All baselines use the same primary-object split, prediction-prefix generation protocol, target labels, evaluation metrics, and available event/object attributes when supported by the corresponding model.

\subsubsection{Evaluation Metrics}
We use Accuracy and macro F1-score as the main evaluation metrics \cite{pasquadibisceglie2022multi}. Accuracy measures overall prediction correctness, while macro F1-score assigns equal weight to each activity label and is therefore more suitable for long-tailed activity distributions. For brevity, F1-score in the tables refers to macro F1-score. All experiments are repeated over five random seeds $\{42,12345,3423,523,6556\}$, and the tables report the mean and sample standard deviation over the five runs.

\subsubsection{Implementation Details}
Each dataset is split by primary-object identifier to avoid assigning prefixes from the same primary object to different partitions. The primary object types are set to \textit{items} for OTC, \textit{application} for BPI~2017 and Intermediate, and \textit{purchase\_order} for P2P. We reserve 20\% of primary objects as the test partition and 10\% of the remaining primary objects as validation, yielding an effective 72:8:20 train/validation/test split. 

Event and object attributes are encoded as input features. Categorical attributes are one-hot encoded using vocabularies built from the training partition only, with missing values mapped to an \textit{unknown} token. Missing numerical values are first filled with zero, and the MinMaxScaler is fitted on the training partition only before being applied to the validation and test partitions. Timestamps are parsed as provided while preserving the original precision and timezone information available in each log. Event ordering and prefix cutoffs are computed on the preserved timestamps, and cyclic sine/cosine features are computed from weekday, hour, minute, and second components. Prefix construction uses a timestamp-based cutoff, so context events must occur earlier than the last observed lifecycle event and the target event is excluded. Lifecycle sequences use a maximum recent window size of 5 and are padded with zeros and boolean masks. 

EHHN is implemented in PyTorch with DHG and trained on a single NVIDIA RTX 3090 GPU. We use AdamW with learning rate $1\mathrm{e}{-4}$, betas $(0.9,0.999)$, no weight decay, and batch size 256. The learning rate follows cosine annealing with a 10-epoch warmup. The maximum number of epochs is 200, with early stopping patience of 20 epochs after a minimum of 20 epochs. The embedding dimension is $D=256$. The micro-spatial stream uses one JEST module, one LCSE module, and two HIE layers with dropout 0.1. The macro-evolution stream uses two TASE-Transformer layers with four attention heads. The classification loss uses label smoothing 0.1, and gradients are clipped at 0.5. 
The auxiliary time-prediction loss uses Smooth L1 loss, with its weight linearly warmed up from 0 to 0.01 over the first 10 epochs. The prototype-diversity loss is enabled for the full EHHN model and uses $\lambda_2=1\mathrm{e}{-3}$. The global prototype counts are selected on the validation partition and set to $K=8,4,8,64$ for OTC, BPI~2017, Intermediate, and P2P, respectively, before final test evaluation.

\subsection{Overall Effectiveness (RQ1)}\label{subsec:rq1}

\begin{table*}[t]
\centering
\caption{Overall comparison with flattening/adapted classical baselines and native OCEL graph baselines (mean$\pm$std over 5 runs). Best per column in \textbf{bold}, second-best underlined.}
\label{tab:overall_results}
\footnotesize
\setlength{\tabcolsep}{2.4pt}
\renewcommand{\arraystretch}{1.1}
\begin{tabular}{l cc cc cc cc}
\toprule
& \multicolumn{2}{c}{OTC} 
& \multicolumn{2}{c}{BPI~2017} 
& \multicolumn{2}{c}{Intermediate} 
& \multicolumn{2}{c}{P2P} \\
\cmidrule(lr){2-3}\cmidrule(lr){4-5}\cmidrule(lr){6-7}\cmidrule(lr){8-9}
Method & Accuracy & F1-score & Accuracy & F1-score & Accuracy & F1-score & Accuracy & F1-score \\
\midrule
\multicolumn{9}{l}{\emph{Flattening/adapted classical baselines}} \\
Dual-View & \underline{0.802}\std{0.004} & \underline{0.685}\std{0.006} & 0.768\std{0.001} & 0.552\std{0.012} & 0.697\std{0.005} & 0.537\std{0.005} & 0.755\std{0.011} & 0.690\std{0.027} \\
PROPHET & 0.754\std{0.038} & 0.560\std{0.035} & 0.762\std{0.041} & 0.579\std{0.031} & 0.595\std{0.067} & 0.417\std{0.067} & 0.797\std{0.002} & \underline{0.842}\std{0.002} \\
Flat-LSTM & 0.773\std{0.003} & 0.660\std{0.015} & 0.758\std{0.001} & 0.561\std{0.003} & \underline{0.703}\std{0.003} & \underline{0.542}\std{0.005} & \underline{0.824}\std{0.019} & 0.816\std{0.009} \\
GAT+LSTM & 0.711\std{0.037} & 0.615\std{0.030} & 0.748\std{0.049} & \underline{0.687}\std{0.067} & 0.510\std{0.067} & 0.436\std{0.084} & 0.773\std{0.003} & 0.766\std{0.004} \\
\midrule
\multicolumn{9}{l}{\emph{Native OCEL graph baselines}} \\
Logistic Regression & 0.293\std{0.017} & 0.090\std{0.004} & 0.632\std{0.001} & 0.495\std{0.009} & 0.556\std{0.008} & 0.455\std{0.008} & 0.652\std{0.022} & 0.571\std{0.019} \\
HOEG & 0.622\std{0.001} & 0.452\std{0.003} & \underline{0.801}\std{0.009} & 0.652\std{0.002} & 0.548\std{0.008} & 0.400\std{0.008} & 0.795\std{0.012} & 0.806\std{0.005} \\
GCN & 0.372\std{0.001} & 0.049\std{0.001} & 0.436\std{0.015} & 0.249\std{0.035} & 0.403\std{0.008} & 0.260\std{0.015} & 0.571\std{0.019} & 0.344\std{0.053} \\
GraphTransformer & 0.371\std{0.001} & 0.056\std{0.003} & 0.593\std{0.006} & 0.357\std{0.028} & 0.574\std{0.010} & 0.408\std{0.019} & 0.719\std{0.013} & 0.573\std{0.043} \\
GAT & 0.371\std{0.001} & 0.056\std{0.003} & 0.623\std{0.002} & 0.462\std{0.010} & 0.541\std{0.012} & 0.432\std{0.024} & 0.617\std{0.013} & 0.521\std{0.021} \\
\midrule
\textbf{EHHN} & \textbf{0.858\std{0.002}} & \textbf{0.809\std{0.003}} & \textbf{0.836\std{0.001}} & \textbf{0.702\std{0.001}} & \textbf{0.719\std{0.007}} & \textbf{0.573\std{0.011}} & \textbf{0.905\std{0.010}} & \textbf{0.903\std{0.014}} \\
\bottomrule
\end{tabular}
\end{table*}

Table~\ref{tab:overall_results} compares EHHN with nine baselines, including flattening/adapted classical methods and native OCEL graph-based methods. EHHN achieves the highest Accuracy and macro F1-score on all four benchmarks among all compared methods. For Accuracy, the strongest baseline methods are Dual-View on OTC, HOEG on BPI~2017, and Flat-LSTM on Intermediate and P2P. EHHN improves over them by $+5.6$, $+3.5$, $+1.6$, and $+8.1$ percentage points, respectively. For macro F1-score, the strongest baseline methods are Dual-View on OTC, GAT+LSTM on BPI~2017, Flat-LSTM on Intermediate, and PROPHET on P2P. EHHN improves over them by $+12.4$, $+1.5$, $+3.1$, and $+6.1$ percentage points. These results indicate that EHHN consistently outperforms the nine baselines across the four evaluated datasets.

The gains are consistent with the dataset complexity profile in Table~\ref{tab:datasets}. On OTC, many objects per event, high event-size entropy, and wide non-primary object branching pose challenges to pairwise graph baselines, while EHHN retains multi-object participation through hyperedges. On BPI~2017, the large scale and long global event-order gap indicate a long-range temporal challenge, where EHHN improves over HOEG, suggesting the benefit of combining structural connectivity with time-aware lifecycle modeling. On Intermediate, whose complexity profile is more balanced, EHHN still provides clear gains over Flat-LSTM. On P2P, the high object-type co-occurrence entropy and low primary-object participation indicate diverse cross-perspective interactions, where EHHN achieves its largest Accuracy gain. Overall, the results suggest that EHHN is particularly useful when structural, temporal, or object-perspective complexity is high, while still maintaining gains on the more balanced benchmark.

\subsection{Ablation Study (RQ2)}\label{subsec:rq2}

Table~\ref{tab:ablation} reports seven ablation variants grouped into three categories. \emph{Operator replacement} replaces the proposed JEST+LCSE+HIE stack with generic \textsc{UniGATConv} or \textsc{HGNNConv} operators. \emph{Stream isolation} keeps only the micro-spatial stream (\textsc{Spatial}) or only the macro-evolution stream (\textsc{Trans}). \emph{Macro-component coupling} tests \textsc{Sp+TASE}, \textsc{Sp+Proto}, and \textsc{Sp+HGNN}, which combine the spatial backbone with the TASE representation, the prototype memory, and an HGNN-based macro alternative, respectively.

The ablation results reveal three main observations. First, replacing the proposed JEST+LCSE+HIE stack with generic \textsc{UniGATConv} or \textsc{HGNNConv} operators reduces performance on all datasets. The gap is especially visible for \textsc{HGNNConv} on Intermediate and P2P, suggesting that event-driven object-state updates and lifecycle-aware refinement offer a useful inductive bias beyond generic hypergraph message passing. Second, in stream isolation, \textsc{Spatial} remains close to the full model, while \textsc{Trans} alone drops by 9.4, 1.4, 1.8, and 7.7 Accuracy points on OTC, BPI~2017, Intermediate, and P2P, respectively. This indicates that the micro-spatial stream is the stronger standalone component, whereas the macro-evolution stream mainly complements it in the full model. Third, in macro-component coupling, \textsc{Sp+TASE} stays close to \textsc{Spatial}, \textsc{Sp+Proto} does not consistently improve over \textsc{Spatial}, and \textsc{Sp+HGNN} is weaker than the full model on all datasets. By contrast, the full model improves F1-score over \textsc{Spatial} by 0.9, 0.6, 1.9, and 0.7 points across the four datasets, suggesting that the temporal and prototype components provide modest but consistent joint gains.

\begin{table*}[t]
\centering
\caption{Ablation study results (mean$\pm$std over 5 runs). Best per column in \textbf{bold}.}
\label{tab:ablation}
\footnotesize
\setlength{\tabcolsep}{3pt}
\renewcommand{\arraystretch}{1.1}
\begin{tabular}{l cc cc cc cc}
\toprule
& \multicolumn{2}{c}{OTC} & \multicolumn{2}{c}{BPI~2017} & \multicolumn{2}{c}{Intermediate} & \multicolumn{2}{c}{P2P} \\
\cmidrule(lr){2-3}\cmidrule(lr){4-5}\cmidrule(lr){6-7}\cmidrule(lr){8-9}
Variant & Accuracy & F1-score & Accuracy & F1-score & Accuracy & F1-score & Accuracy & F1-score \\
\midrule
\textbf{EHHN} & \textbf{0.858\std{0.002}} & \textbf{0.809\std{0.003}} & \textbf{0.836\std{0.001}} & \textbf{0.702\std{0.001}} & \textbf{0.719\std{0.007}} & \textbf{0.573\std{0.011}} & \textbf{0.905\std{0.010}} & \textbf{0.903\std{0.014}} \\
\midrule
\multicolumn{9}{l}{\emph{(i) Operator replacement}} \\
\textsc{UniGATConv} & 0.846\std{0.000} & 0.788\std{0.003} & 0.830\std{0.001} & 0.693\std{0.001} & 0.705\std{0.002} & 0.556\std{0.006} & 0.886\std{0.004} & 0.881\std{0.005} \\
\textsc{HGNNConv}   & 0.844\std{0.004} & 0.782\std{0.014} & 0.828\std{0.001} & 0.690\std{0.003} & 0.681\std{0.005} & 0.533\std{0.009} & 0.825\std{0.008} & 0.796\std{0.016} \\
\midrule
\multicolumn{9}{l}{\emph{(ii) Stream isolation}} \\
\textsc{Spatial} & 0.852\std{0.001} & 0.800\std{0.001} & 0.834\std{0.001} & 0.696\std{0.003} & 0.705\std{0.006} & 0.554\std{0.007} & 0.901\std{0.011} & 0.896\std{0.013} \\
\textsc{Trans}   & 0.764\std{0.001} & 0.599\std{0.003} & 0.822\std{0.001} & 0.666\std{0.001} & 0.701\std{0.004} & 0.535\std{0.009} & 0.828\std{0.012} & 0.804\std{0.022} \\
\midrule
\multicolumn{9}{l}{\emph{(iii) Macro-component coupling}} \\
\textsc{Sp+TASE} & 0.852\std{0.002} & 0.801\std{0.004} & 0.836\std{0.001} & 0.700\std{0.001} & 0.712\std{0.006} & 0.561\std{0.010} & 0.902\std{0.017} & 0.899\std{0.024} \\
\textsc{Sp+Proto}      & 0.845\std{0.002} & 0.794\std{0.001} & 0.834\std{0.001} & 0.697\std{0.002} & 0.703\std{0.006} & 0.564\std{0.012} & 0.893\std{0.014} & 0.885\std{0.016} \\
\textsc{Sp+HGNN}       & 0.830\std{0.002} & 0.780\std{0.001} & 0.824\std{0.001} & 0.680\std{0.003} & 0.619\std{0.004} & 0.477\std{0.013} & 0.852\std{0.008} & 0.845\std{0.008} \\
\bottomrule
\end{tabular}
\end{table*}

\begin{figure*}[t]
\centering
\begin{subfigure}[b]{0.98\columnwidth}
\centering
\includegraphics[width=\linewidth]{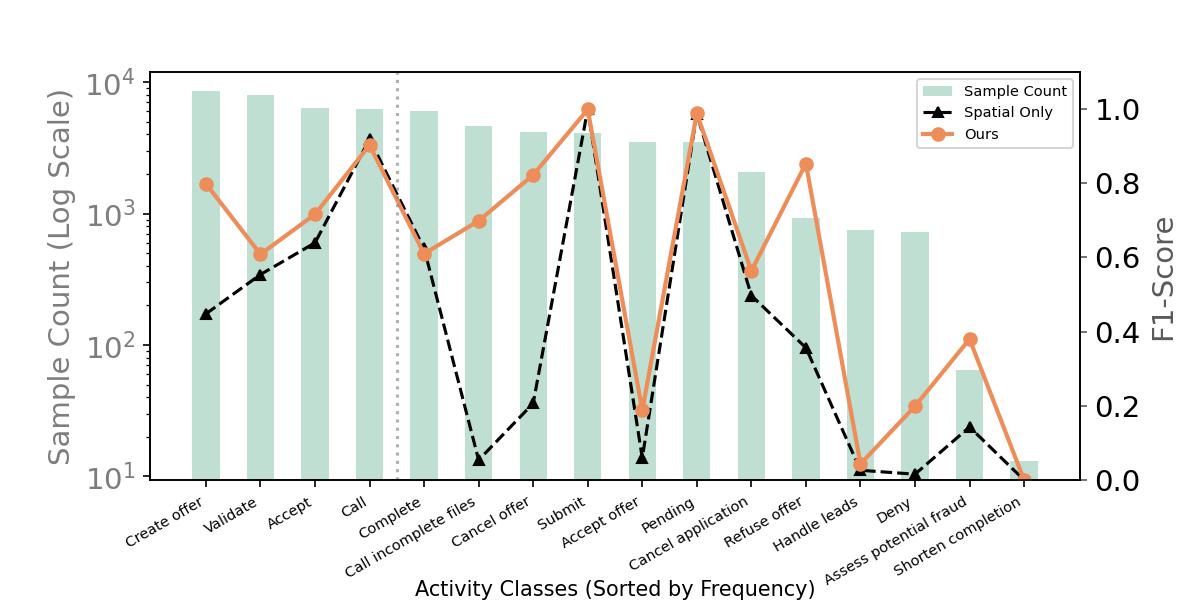}
\caption{BPI~2017}
\label{fig:longtail_bpi}
\end{subfigure}
\hfill
\begin{subfigure}[b]{0.98\columnwidth}
\centering
\includegraphics[width=\linewidth]{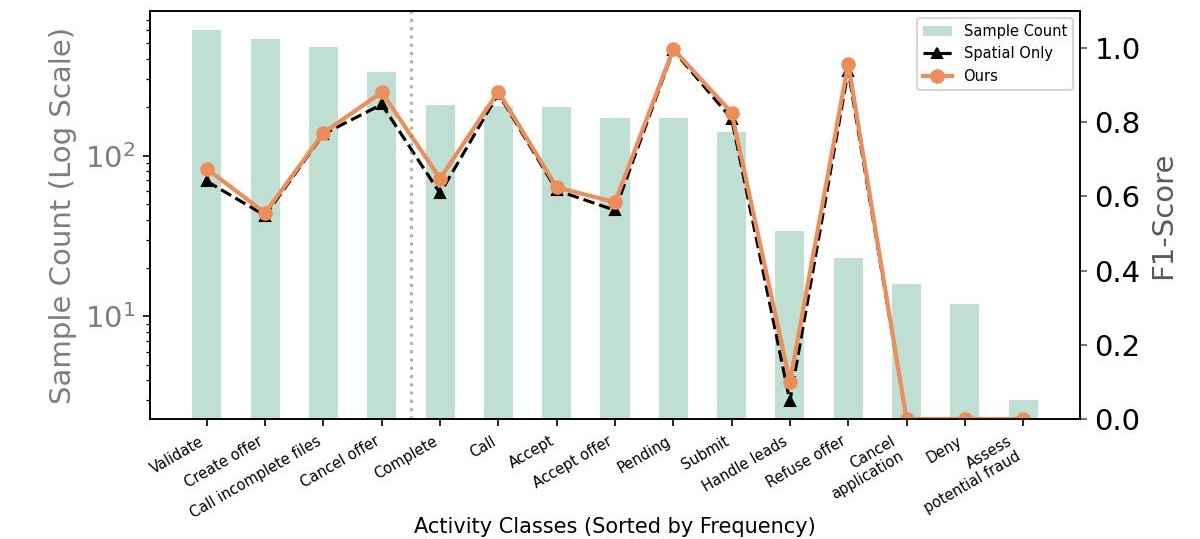}
\caption{Intermediate}
\label{fig:longtail_intermediate}
\end{subfigure}
\vspace{-2mm}
\caption{Frequency-stratified per-class F1-score on BPI~2017 and Intermediate.}
\label{fig:longtail}
\end{figure*}

Because the macro-evolution stream brings only modest gains in aggregate metrics, we further examine whether these gains vary across activity classes. We focus on BPI~2017 and Intermediate since they contain enough activity classes to make frequency-stratified per-class comparison meaningful. Fig.~\ref{fig:longtail} reports frequency-stratified per-class F1-score, where classes are sorted from frequent to rare. On BPI~2017, EHHN and \textsc{Spatial} perform similarly on frequent classes, while EHHN shows clearer gains in several mid-frequency and tail classes, such as \textit{Call incomplete files}, \textit{Cancel offer}, and \textit{Refuse offer}. On Intermediate, the two curves are mostly close across activity classes, indicating that the additional macro-evolution information contributes less to class-level separation on this dataset. Overall, the analysis suggests that the macro-evolution stream provides class-dependent refinements rather than uniformly large gains across all datasets.

\subsection{Efficiency and Deployment Feasibility (RQ3)}\label{subsec:rq3}

\begin{table}[t]
\centering
\caption{Training and inference efficiency comparison. Bold indicates the better value between the two structure-preserving methods, EHHN and HOEG, for training time and memory.}
\label{tab:efficiency}
\footnotesize
\setlength{\tabcolsep}{3pt}
\renewcommand{\arraystretch}{1.05}
\resizebox{\columnwidth}{!}{%
\begin{tabular}{llccc}
\toprule
Dataset & Metric & EHHN & Flat-LSTM & HOEG \\
\midrule
\multirow{3}{*}{OTC}
& Train time (s/epoch) & 75.17 & 1.41 & \textbf{56.03} \\
& Latency (ms/sample)  & 10.66 & 0.72 & 5.64 \\
& Memory (GB)          & 0.99  & 0.05 & \textbf{0.81} \\
\midrule
\multirow{3}{*}{BPI~2017}
& Train time (s/epoch) & \textbf{133.11} & 7.89 & 671.85 \\
& Latency (ms/sample)  & 10.15 & 0.75 & 3.37 \\
& Memory (GB)          & \textbf{0.19} & 0.05 & 4.60 \\
\midrule
\multirow{3}{*}{Interm.}
& Train time (s/epoch) & 9.43 & 0.54 & \textbf{4.23} \\
& Latency (ms/sample)  & 35.80 & 1.02 & 3.32 \\
& Memory (GB)          & \textbf{0.23} & 0.06 & \textbf{0.23} \\
\midrule
\multirow{3}{*}{P2P}
& Train time (s/epoch) & 2.91 & 0.13 & \textbf{2.48} \\
& Latency (ms/sample)  & 37.41 & 1.08 & 5.40 \\
& Memory (GB)          & \textbf{0.23} & 0.05 & 0.24 \\
\bottomrule
\end{tabular}}
\end{table}

Table~\ref{tab:efficiency} compares EHHN with two baselines that represent different efficiency-structure trade-offs. Flat-LSTM is a flattened sequence model that illustrates the efficiency of ignoring the native multi-object OCEL structure. HOEG is a native OCEL graph baseline and is used as the main structure-preserving comparison for training cost, latency, and GPU memory. On OTC, EHHN is slower and uses slightly more memory than HOEG, indicating that its hypergraph construction and dual-stream encoding introduce extra overhead on this dataset. On BPI~2017, the largest benchmark, EHHN shows its clearest scalability advantage over HOEG: training time decreases from 671.85 to 133.11 s/epoch, and peak GPU memory decreases from 4.60 to 0.19~GB, corresponding to a $24\times$ reduction. On Intermediate, EHHN and HOEG use the same peak memory, while HOEG is faster. On P2P, their training time and memory are close, although EHHN has higher inference latency.

Overall, EHHN does not aim to be the fastest predictor. Flat-LSTM remains much faster because it uses a flattened sequence representation, and HOEG often has lower per-sample latency. The practical value of EHHN is that it improves prediction quality while keeping GPU memory manageable for structure-preserving OCEL prediction, especially on the largest benchmark, at the cost of higher inference latency.

\subsection{Sensitivity Analysis of Prototype Count $K$ (RQ4)}\label{subsec:rq4}

\begin{figure*}[htp]
\centering
\begin{subfigure}[t]{0.43\textwidth}\centering
\includegraphics[width=\textwidth]{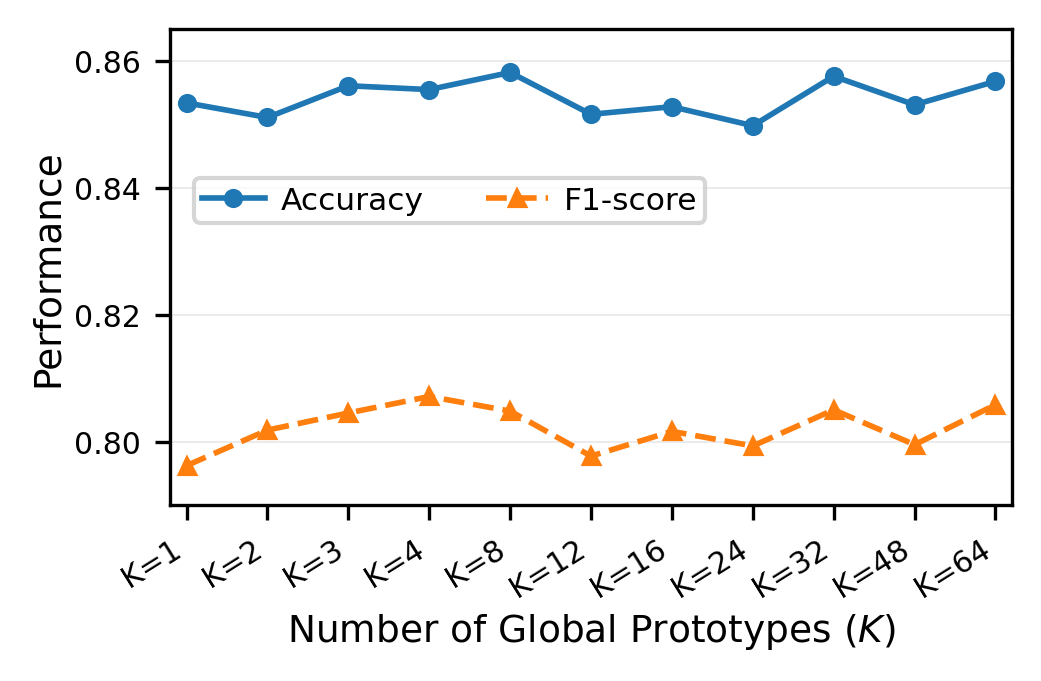}
\caption{OTC}\label{fig:k_otc}\end{subfigure}
\hfill
\begin{subfigure}[t]{0.43\textwidth}\centering
\includegraphics[width=\textwidth]{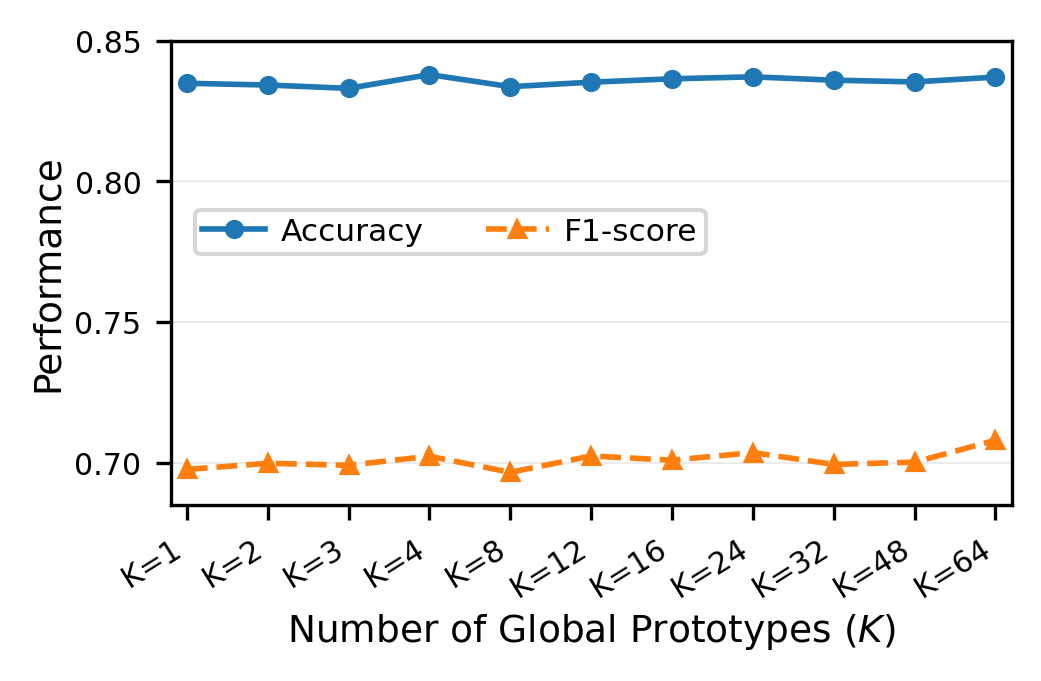}
\caption{BPI~2017}\label{fig:k_bpi}\end{subfigure}
\vspace{0.4em}
\begin{subfigure}[t]{0.43\textwidth}\centering
\includegraphics[width=\textwidth]{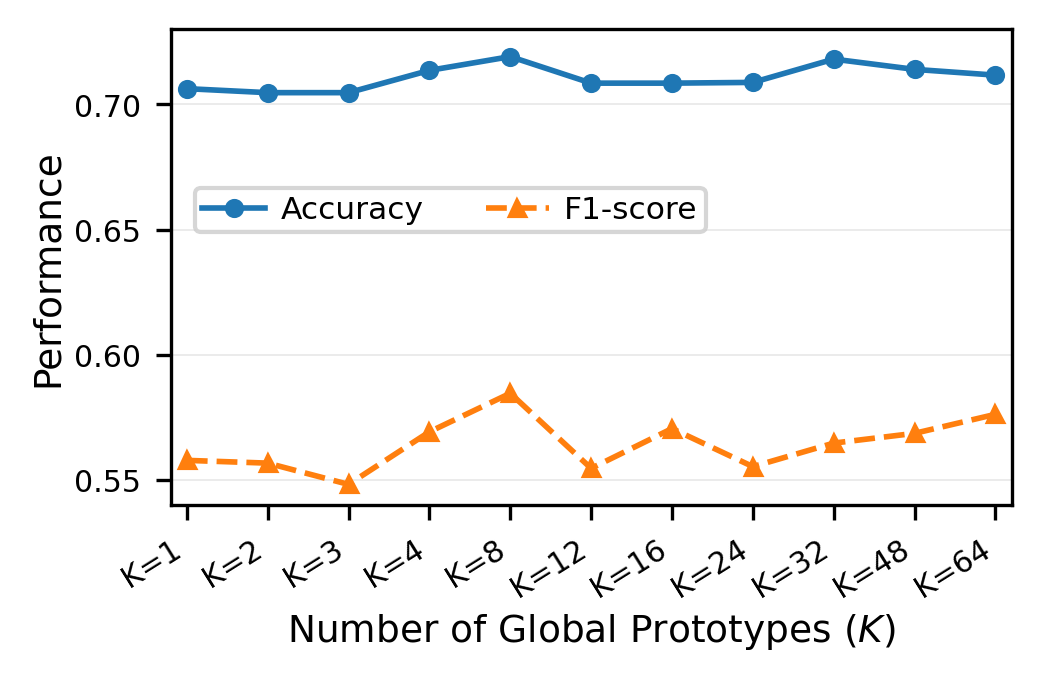}
\caption{Intermediate}\label{fig:k_inter}\end{subfigure}
\hfill
\begin{subfigure}[t]{0.43\textwidth}\centering
\includegraphics[width=\textwidth]{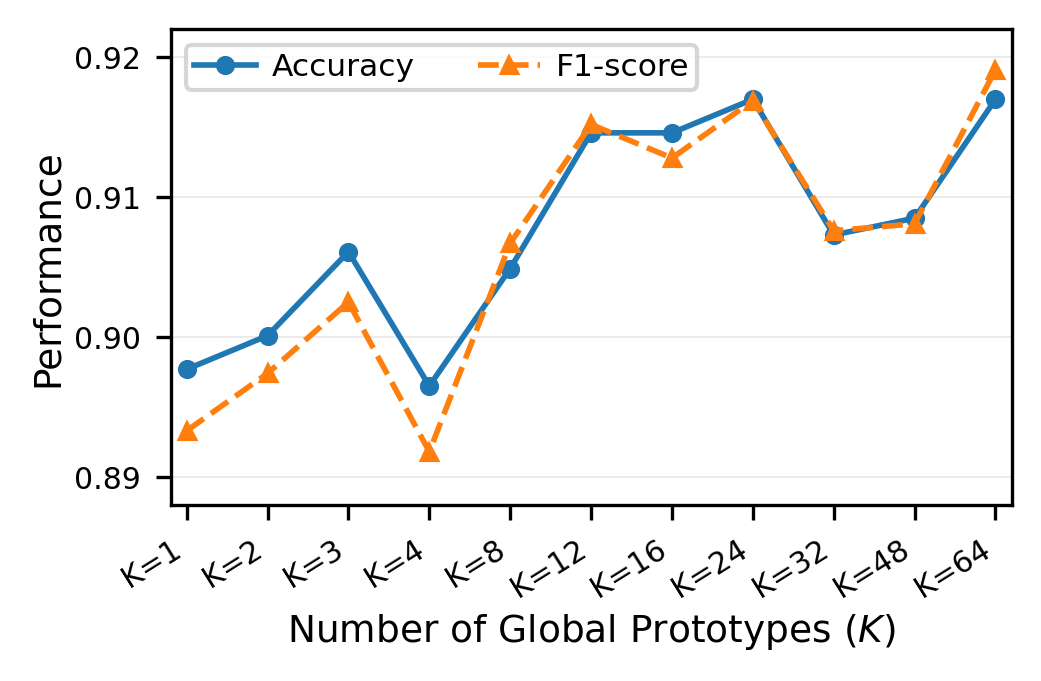}
\caption{P2P}\label{fig:k_p2p}\end{subfigure}
\caption{Sensitivity of EHHN to prototype count $K$ on four benchmarks.}
\label{fig:k_analysis}
\end{figure*}

The macro-evolution stream uses a global prototype memory with $K$ learnable prototypes to represent global execution patterns. Fig.~\ref{fig:k_analysis} reports a sensitivity analysis over $K\in\{1,2,3,4,8,12,16,24,32,48,64\}$ to examine the robustness of EHHN to the prototype count. This analysis is not used to select the hyperparameters for the final test results. Overall, EHHN is not highly sensitive to $K$: Accuracy and macro F1-score fluctuate within a relatively small range across most settings, indicating that the prototype memory provides stable guidance without requiring a finely tuned prototype count. The curves show dataset-dependent preferences: BPI~2017 performs well with a compact prototype memory, OTC and Intermediate favor moderate values, and P2P benefits from a larger memory.

The dataset-dependent best $K$ values can be interpreted through the complexity profile in Table~\ref{tab:datasets}. Since prototypes are intended to represent recurring global execution patterns, a larger $K$ is more useful when the log contains more diverse object-perspective interactions. Such diversity is reflected by high object-type co-occurrence entropy and low primary-object participation: events involve more varied combinations of object types, and the primary-object perspective is less dominant. This explains why P2P, which has the strongest object-perspective complexity, reaches its best macro F1-score at $K=64$. In contrast, BPI~2017 is the largest benchmark and has the largest global event-order gap, but its object-perspective complexity is not the highest, so a small prototype memory can already cover its dominant execution patterns. OTC is the most structurally complex benchmark, but its object-type combinations are less diverse than P2P and its primary-object participation is higher, making a moderate $K$ sufficient. Intermediate has a relatively balanced profile and is also well served by a moderate $K$.

\subsection{Interpretability and Robustness (RQ5)}\label{subsec:further}

\subsubsection{Prototype Interpretability}
\begin{figure}[t]
\centering
\includegraphics[width=0.9\linewidth]{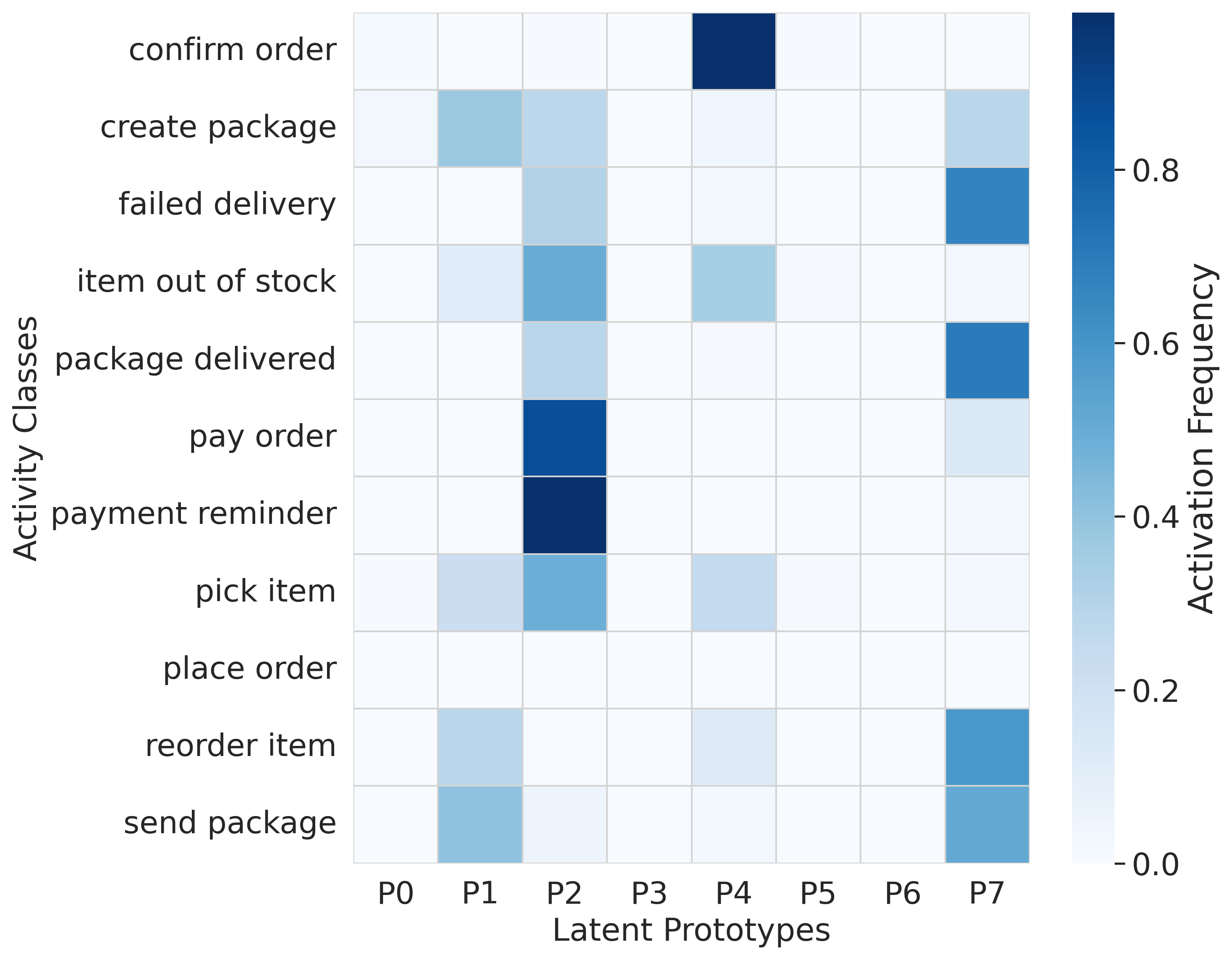}
\caption{Prototype activation heatmap on OTC.}
\label{fig:prototype_heatmap}
\end{figure}

In EHHN, the macro-evolution stream uses a global prototype memory with $K$ learnable prototypes to capture dataset-level execution patterns. To examine whether these prototypes are interpretable, we visualize prototype activations on OTC. We choose OTC because its activity labels have clear business meanings, including payment, packaging, delivery, and exception handling. For OTC, the selected prototype count is $K=8$, so the columns $P_0$--$P_7$ in Fig.~\ref{fig:prototype_heatmap} correspond to the eight learned prototypes. Each cell records how frequently prefixes ending with a given activity activate a prototype, with darker colors indicating higher activation frequency. The heatmap shows domain-related grouping without manually defined process modes: $P_2$ is strongly activated by \textit{pay order} and \textit{payment reminder}, corresponding to a financial-transaction pattern, while $P_7$ is activated by \textit{send package}, \textit{package delivered}, \textit{failed delivery}, and \textit{reorder item}, corresponding to a logistics and exception-handling pattern. This suggests that the prototype memory learns interpretable execution patterns.

\subsubsection{Robustness to Noisy Logs}
Real-world event logs may contain corrupted attributes or event features. Table~\ref{tab:noise} reports Accuracy and macro F1-score under two 30\% noise settings: \emph{attribute masking}, which randomly masks node attributes, and \emph{event-feature masking}, which randomly masks event features. We compare with Flat-LSTM because it is a flattened sequence baseline without explicit object-centric structural redundancy, making it a suitable contrast for evaluating whether EHHN's hypergraph structure improves robustness.

EHHN remains more stable than Flat-LSTM under both noise types. On OTC, EHHN loses at most 4.9 Accuracy points and 6.3 F1 points, whereas Flat-LSTM loses up to 32.6 Accuracy points and 30.2 F1 points. On Intermediate, EHHN loses at most 5.2 Accuracy points and 5.7 F1-score points, while Flat-LSTM loses up to 12.9 Accuracy points and 10.7 F1-score points. These results suggest that the hypergraph representation provides useful structural redundancy: when local attributes or event features are corrupted, neighboring event-object context and lifecycle information can still support prediction. In contrast, flattened sequence models have less object-centric context to rely on once event or attribute features are corrupted.

\begin{table}[t]
\centering
\caption{Robustness under 30\% noisy event logs.}
\label{tab:noise}
\footnotesize
\setlength{\tabcolsep}{2.6pt}
\renewcommand{\arraystretch}{1.08}
\begin{tabular}{llcccc}
\toprule
Dataset & Noise setting & \multicolumn{2}{c}{EHHN} & \multicolumn{2}{c}{Flat-LSTM} \\
\cmidrule(lr){3-4}\cmidrule(lr){5-6}
 & & Accuracy & F1 & Accuracy & F1 \\
\midrule
\multirow{3}{*}{OTC}
& No noise & 0.858 & 0.809 & 0.773 & 0.660\\
& Attribute masking & 0.810 & 0.751 & 0.447 & 0.358 \\
& Event-feature mask & 0.809 & 0.746 & 0.539 & 0.447 \\
\midrule
\multirow{3}{*}{Intermediate}
& No noise & 0.719 & 0.573 & 0.702 & 0.543 \\
& Attribute masking & 0.667 & 0.516 & 0.573 & 0.441 \\
& Event-feature mask & 0.687 & 0.532 & 0.573 & 0.436 \\
\bottomrule
\end{tabular}
\end{table}

\section{Conclusion}\label{sec:conclusion}

This paper addressed object-centric next activity prediction by modeling prediction prefixes extracted from OCELs as heterogeneous hypergraphs and encoding them with a dual-stream architecture. The hypergraph representation retains multi-object event participation within a bounded prefix without flattening the log or decomposing retained event-object relations into pairwise links. The dual-stream architecture connects local event-driven object-state evolution with time-aware trajectory modeling and prototype-based global guidance. Experiments on four public OCEL benchmarks show that EHHN achieves higher Accuracy and macro F1-score than nine baselines, with improvements of up to 8.1 and 12.4 percentage points, respectively. The efficiency results show that EHHN keeps GPU memory practical for structure-preserving prediction, especially on the largest benchmark. The interpretability and robustness analyses suggest that the learned prototypes capture meaningful process modes and that the hypergraph representation is more resilient than flattened sequence modeling under noisy attributes and event features. Future work will explore adaptive primary-object selection, online updating for streaming OCELs, and validation on industrial service logs.

\bibliographystyle{IEEEtran}
\bibliography{main}

\end{document}